\documentclass{svproc}

\pdfoutput=1
\usepackage{url}

\usepackage{graphicx}
\usepackage{subcaption}
\usepackage{hyperref}
\usepackage{amsmath}
\usepackage{cite}
\usepackage{bm}
\usepackage{xcolor}

\DeclareMathOperator{\atantwo}{atan2}

\begin{document}

\mainmatter              % start of a contribution
\title{Towards Efficient Full Pose Omnidirectionality with Overactuated MAVs}
\titlerunning{Towards Efficient Full Pose Omnidirectionality with Overactuated MAVs}
\author{Karen Bodie \and Zachary Taylor \and Mina Kamel \and Roland Siegwart}
\authorrunning{Karen Bodie et al.} % abbreviated author list (for running head)
\tocauthor{Karen Bodie, Zachary Taylor, Mina Kamel, and Roland Siegwart}
\institute{Autonomous Systems Lab, ETH Z\"{u}rich, Switzerland}

\maketitle              % typeset the title of the contribution

\begin{abstract}
 Omnidirectional MAVs are a growing field, with demonstrated advantages for aerial interaction and uninhibited observation.
 While systems with complete pose omnidirectionality and high hover efficiency have been developed independently, a robust system that combines the two has not been demonstrated to date.
 This paper presents VoliroX: a novel omnidirectional vehicle that can exert a wrench in any orientation while maintaining efficient flight configurations.
 The system design is presented, and a 6 DOF geometric control that is robust to singularities.
 Flight experiments further demonstrate and verify its capabilities.

\end{abstract}

\section{Introduction}
Traditional rotary micro aerial vehicles (MAVs) are underactuated, providing only 4 controllable degrees of freedom (DOFs) by nature of their aligned propeller axes.
This can lead to unstable behavior when a force and torque are applied to such a system in flight.
Omnidirectional MAVs allow for decoupling of the translational and rotational dynamics, permitting stable interaction with the environment in any pose.
With the addition of complete pose omnidirectionality, the system can achieve uninhibited aerial movement and robust tracking of 6 DOF trajectories, a unique boon for aerial filming and 3D mapping, as well as configuration-based navigation in constrained environments.

A dominant struggle in the development of MAVs is achieving a good compromise between performance and efficiency.
For aerial interaction and high dynamic control authority in omnidirectional flight, performance can be represented by the force and torque control volumes of a system.
For pose-omnidirectional platforms, the force envelope must exceed gravity in all directions with an additional buffer to maintain dynamic movement.
Countering this performance goal is the desire for high efficiency and longer flight times, which is compromised in systems that generate high internal forces (thrust forces which counteract each other), or add additional weight for actuation.

\paragraph{Problem Statement:}
This paper presents a novel platform that achieves both force- and pose-omnidirectionality with highly dynamic capabilities, while maintaining high efficiency in hover, by nature of its six actuated tilt arms.
Experimental results demonstrate that the additional complexity and weight of the tilt-arm configuration are justified by the system's performance.

\paragraph{Related Work:}
Within the past 5 years, huge growth has occurred in the field of fully actuated omnidirectional MAVs.
These works can be generally categorized as fixedly tilted rotor platforms, and tilt-arm rotor unit platforms.
By its design, a platform with fixedly tilted rotors that is able to generate a significant wrench on the environment creates a proportionately significant amount of internal force, which directly detracts from flight efficiency \cite{brescianini2016design, park2018odar, park2016design, staub2018towards}.

Tilt-arm platforms can achieve optimal hover efficiency in the absence of external disturbances, when all propeller thrust vectors are aligned against gravity.
This has been demonstrated in the form of a quadrotor \cite{falconi2012dynamic, ryll2015novel} with limited roll and pitch,
and also in the form of a hexarotor, with our platform's predecessor Voliro \cite{kamel2018voliro},
which is limited by low thrust and singularity issues in certain configurations.
These platforms achieve force omnidirectionality with high hover efficiency at the cost of additional complexity, weight, and inertia.
An intersection of the two concepts which couples the tilt axes to a single motor has also been evaluated in simulation \cite{ryll2016modeling},
enabling efficient flight, but without pose omnidirectionality.

%%%%%%%%%%%%%%%%%%%%%%%%%%%%%%%%%%
\section{Technical Approach}
%%%%%%%%%%%%%%%%%%%%%%%%%%%%%%%%%%

\subsection{System Design}

\begin{figure}
 \centering
\includegraphics[width=\textwidth]{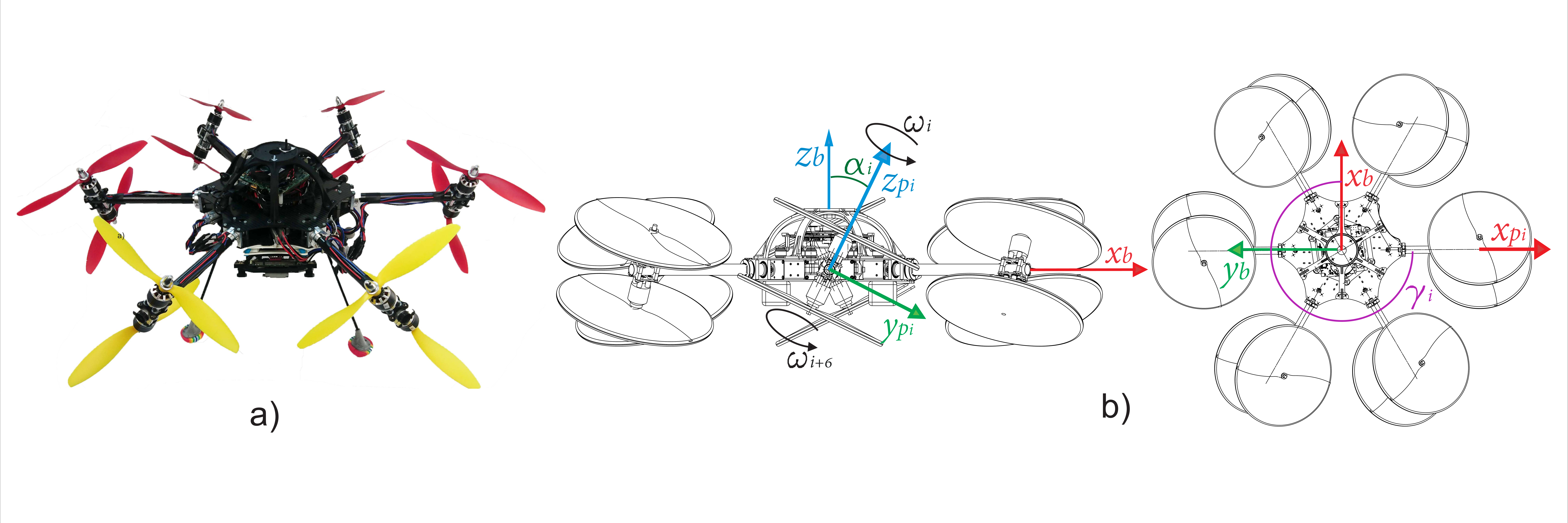}
 \caption{(a) Prototype system, (b) coordinate frames and variables}
 \label{fig:description}
\end{figure}

This paper presents the VoliroX platform: a 12 rotor MAV with 6 tiltable arms, shown in \autoref{fig:description}(a).
The system layout consists of equally spaced arms along the $z_b$-plane, with two rotors per arm to balance the rotational inertia.
Two KDE2315XF-885 motors per arm with 9in propellers also provide high thrust (13.7N per motor), enabling dynamic flight even in the least efficient configurations.
The arm is tilted by a Dynamixel XL430 servo actuator, located in the base to reduce system inertia.
Upper and lower propellers counter-rotate such that drag torque from the propellers is canceled in the nominal case, and can be controlled by varying the relative inputs to the rotors.
Moreover, the counter-rotating propeller configuration minimizes the gyroscopic moment on the tilting mechanism, allowing the use of a small Dynamixel servo.
An onboard Intel NUC i7 manages higher level computation and communication tasks, while the combined position and attitude controller operates on a PX4 flight controller.
The system is powered by two 3800 mAh 6s LiPo batteries, and can be tethered for the case of near-horizontal endurance flights.
The total system mass is 4kg, and it can generate 150N of force at maximum thrust in the horizontal hover configuration.

Coordinate systems for the platform are described in \autoref{fig:description}(b), with the body-fixed coordinate frame $x_b$, $y_b$, $z_b$, and definitions of the fixed arm spacing angles $\gamma_i$ and tilting angles $\alpha_i$.

%%%%%%%%%%%%%%%%%%%%%%%%%%%%%%%%%%
\subsection{System Dynamics and Control} \label{section:control}
%%%%%%%%%%%%%%%%%%%%%%%%%%%%%%%%%%
The following assumptions are adopted to simplify the model:
\renewcommand{\labelitemi}{$\vcenter{\hbox{\tiny$\bullet$}}$}
\begin{itemize}
 \item The body structure is rigid and symmetric.
 \item Thrust and drag torques are proportional to the square of rotor’s speed, and rotors are able to achieve desired speeds $\omega_i$ with negligible transients.
 \item The primary axes of the system correspond with the principal axes of inertia.
 \item The dynamics of the tilt motors are independent of the rotational speed of rotors.
 \item The thrust and drag torques produced by each rotor are independent, i.e. there is no airflow interference.
\end{itemize}

The system dynamics are derived by the Newton-Euler approach under the stated assumptions,
resulting in the following equations of motion expressed in the body-fixed frame:

\begin{equation} \label{eq:EOM}
 \begin{bmatrix}
  \bm{F}_{b}    \\
  \bm{\tau}_{b}
 \end{bmatrix}
 =
 \begin{bmatrix}
  m \bm{I}_{3} & 0          \\
  0            & \bm{J}_{b}
 \end{bmatrix}
 \begin{bmatrix}
  \ddot{\bm{x}}_{b}     \\
  \dot{\bm{\omega}}_{b}
 \end{bmatrix}
 +
 \begin{bmatrix}
  \bm{\omega}_{b} \times{} m \dot{\bm{x}}_{b}      \\
  \bm{\omega}_{b}\times \bm{J}_{b} \bm{\omega}_{b}
 \end{bmatrix}
\end{equation}

\noindent where $\bm{F}_{b}$ and $\bm{\tau}_{b}$ are the total forces and torques on the system, $\dot{\bm{x}}_{b}$ and $\dot{\bm{x}}_{b}$ are velocity and acceleration of the origin,
and $\bm{\omega}_{b}$ is the angular body velocity, and $m$ and $\bm{J}_{b}$ are the mass and inertia,
where subscript ${}_{b}$ denotes quantities expressed in the body-fixed frame.

Trajectory tracking on SE(3) is used to guarantee local exponential stability \cite{lee2010geometric},
and has been extended to the omnidirectional case \cite{invernizzi2017geometric}.
The control allocation presented here is an extension of \cite{kamel2018voliro, falconi2012dynamic}, a nonlinear function of tilt angles $\alpha_i$.
The force allocation matrix mapping propeller thrust and torques to the net body force and torque is defined as $\bm{A}_\alpha$, and can be thought of as the instantaneous allocation matrix,
since it does not depend on any delay in response of the tilting motors.
One can take advantage of the fact that all entries in $\bm{A}_\alpha$ are linear combinations of $\sin(\alpha_i)$ and $\cos(\alpha_i)$,
and extract them to give a simplified allocation matrix $\bm{A}$.
The result vector $\bm{u}$ has 4 values for each tilt arm, with upper rotors $j\in\{1...6\}$ and lower rotors $\{7...12\}$.
For a system with 6 arms spaced evenly around the $z_b$-axis at rotation angles $\gamma_i$, with arm length $l_x$, rotor spin direction $s_j$, and thrust and drag coefficients $c_f$ and $c_d$:

\begin{equation}
 \begin{bmatrix}\bm{F}_{b} \\ \bm{\tau}_{b}\end{bmatrix}
 = \bm{A}_{\alpha} \bm{\Omega}
 = \bm{A} \bm{u}
 \quad \quad
 \bm{\Omega} = \begin{bmatrix}\Omega_{1} \\ \vdots \\ \Omega_{12} \end{bmatrix} = \begin{bmatrix}\omega_{1}^2 \\ \vdots \\ \omega_{12}^2 \end{bmatrix}
 % \quad
 \label{eq:allocation}
\end{equation}
\begin{equation*}
 \begin{matrix}
  \bm{A} = c_f
  \begin{bmatrix}
  \sin(\gamma_i)            & 0                   & \hdots \\
  -\cos(\gamma_i)           & 0                   & \hdots \\
  0                         & 1                   & \hdots \\
  -s_j c_d \sin(\gamma_i) & l_x \sin(\gamma_i)  & \hdots \\
  s_j c_d \cos(\gamma_i)  & -l_x \cos(\gamma_i) & \hdots \\
  -l_x                      & -s_j c_d            & \hdots
  \end{bmatrix}
                            & \quad               &
  \bm{u} =
  \begin{bmatrix}
  \sin(\alpha_i) \Omega_{i,}  \\
  \cos(\alpha_i) \Omega_{i}  \\
  \vdots \\
  \sin(\alpha_i) \Omega_{i+6} \\
  \cos(\alpha_i) \Omega_{i+6} \\
  \vdots
  \end{bmatrix}
  &
  \begin{matrix}
  \forall i \in \{1...6\} \\
  \forall j \in \{1...12\}
 \end{matrix}
 \end{matrix}
\end{equation*}

\noindent where $\Omega_i$ is the square of the rotor velocity $\omega_i$, and is always chosen to be positive with the proposed allocation method.

A control law for the system is derived from (\ref{eq:allocation}) using the Moore-Penrose pseudo-inverse of the force allocation matrix $\bm{A}$ to compute the actuator commands from a desired 6 DOF wrench $[\bm{F}_d \ \bm{\tau}_d]^\top$ expressed in the body-fixed frame:

\begin{equation}
  \bm{u} = \bm{A}^\dagger \begin{bmatrix}\bm{F}_{d} \\ \bm{\tau}_{d}\end{bmatrix}
\end{equation}

The desired forces and torques on the system are expressed as

\begin{equation}
  \begin{matrix}

\bm{F}_{d} = m (\bm{R}^\top (-k_p \bm{e}_p - k_v \bm{e}_v + \ddot{\bm{x}}_{sp} + \bm{g})
    + (\bm{\omega}_b \times \bm{R}^\top \dot{\bm{x}}))

\\

\bm{\tau}_{d} = \bm{J}_b (-k_R \bm{e}_{R} -
                          k_{\omega} \bm{e}_{\omega}) +
                          (\bm{\omega}_b \times \bm{J}_b \bm{\omega}_b)
                          + (\bm{x}_{com} \times \bm{F}_{d})

\end{matrix}
\label{eq:f_d}
\end{equation}

\noindent and are based on the equations of motion described in (\ref{eq:EOM}), with accelerations taken as a combination of position and velocity errors multiplied by tuned gain values.
To compensate for a center of mass offset $\bm{x}_{com}$, a counter-torque is added to the desired torque equation.
In (\ref{eq:f_d}), $\bm{R} \in \text{SO}(3)$ is the rotation matrix from the body-fixed frame to the inertial frame,
$\bm{g}$ is the gravity vector along the inertial frame $z$-axis, and $k_p$, $k_v$, $k_R$, and $k_{\omega}$, are positive constant gains on position, velocity, rotation, and angular rate errors.
Errors are expressed in the inertial frame relative to set point values with subscript ${}_{sp}$, with angular error terms defined in (\ref{eq:error}) according to \cite{lee2010geometric}.

\begin{equation}
  \begin{matrix}
                          \bm{e}_p = \bm{x} - \bm{x}_{sp}
                          \\
                          \bm{e}_v = \dot{\bm{x}} - \dot{\bm{x}}_{sp}
                          \\
                          \bm{e}_R = \frac{1}{2}(\bm{R}_{sp}^\top \bm{R} - \bm{R}^\top \bm{R}_{sp})^\vee
                          \\
                          \bm{e}_{\omega} = \bm{\omega}_b - \bm{R}^\top \bm{R}_{sp} \bm{\omega}_{sp}

  \end{matrix}
  \label{eq:error}
\end{equation}

To extract $\alpha_i$ values from the result vector $\bm{u}$, each pair of rows is summed, and the trigonomic identity $\theta = \atantwo(\sin(\theta),\cos(\theta))$ is used.
The resulting $\alpha_{i}$ values are then limited according to $\dot{\alpha}_{\max}$.
To extract $\Omega_i$ values, consider the identity $\sin^2(\theta) + \cos^2(\theta) = 1$, to obtain the solution as follows:

\begin{equation}
  \label{eq:omegas}
  \begin{matrix}
  \Omega_i = \Omega_i \sin^2(\alpha_i) + \Omega_i \cos^2(\alpha_i) = \sin(\alpha_i) u_{2i} + \cos(\alpha_i) u_{2i+1} \\
  \Omega_{i+6} = \Omega_{i+6} \sin^2(\alpha_i) + \Omega_i \cos^2(\alpha_i) = \sin(\alpha_i) u_{2(i + 6)} + \cos(\alpha_i) u_{2(i+6)+1}
\end{matrix}
\end{equation}

Notice that the solution in (\ref{eq:omegas}) projects the desired force onto the best achievable tilt angle.
If the direction of desired force would require an inversion of the propeller,
the corresponding $\Omega_i$ is then set to zero, permitting only positive thrust values.

\begin{figure}
 \centering
 \includegraphics[width=\textwidth]{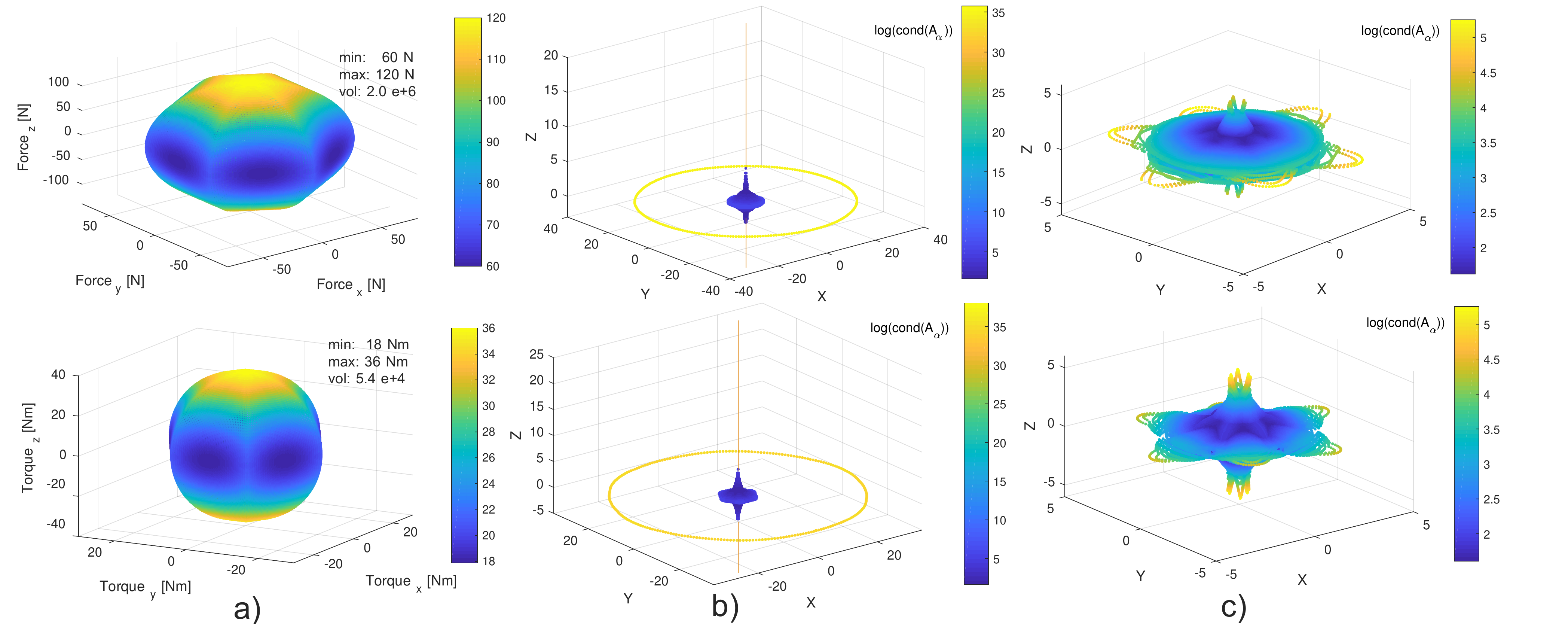}
 \caption{(a) Maximum force and torque envelopes, (b) log of condition numbers of $\bm{A}_\alpha(\bm{\alpha})$, and (b) log of condition number for modified $\bm{A}_\alpha(\tilde{\bm{\alpha}})$, with scale significantly zoomed in.}
 \label{fig:ForceCondition}
\end{figure}

Thrust values for individual motors have been computed based on test results of a double-rotor propeller group on an RCbenchmark test stand, to measure the rotor $z_{p_i}$-axis thrust and torque.
Thrust and torque values are correlated linearly with pulse-width modulation (PWM) commands sent to the rotors' electronic speed controllers (ESCs).
To remain in a thrust region that maintains a good linear approximation, and to reduce the risk of overheating,
the maximum output of the rotors is limited to 80\%,
which corresponds with a thrust of 20 N for each double rotor unit.
Using this test data, maximum achievable force and torque envelopes of the system can be determined by feeding commands forward through the allocation matrix (\ref{eq:allocation}).
The resulting plots for maximum force and torque envelopes are shown in \autoref{fig:ForceCondition}(a).
The force envelope is computed in the absence of torque, and vice versa.

%%%%%%%%%%%%%%%%%%%%%%%%%%%%%%%%%%
\subsection{Singularity Handling}
%%%%%%%%%%%%%%%%%%%%%%%%%%%%%%%%%%

\begin{figure}
 \centering
 \begin{subfigure}[b]{0.80\textwidth}
  \includegraphics[width=\textwidth]{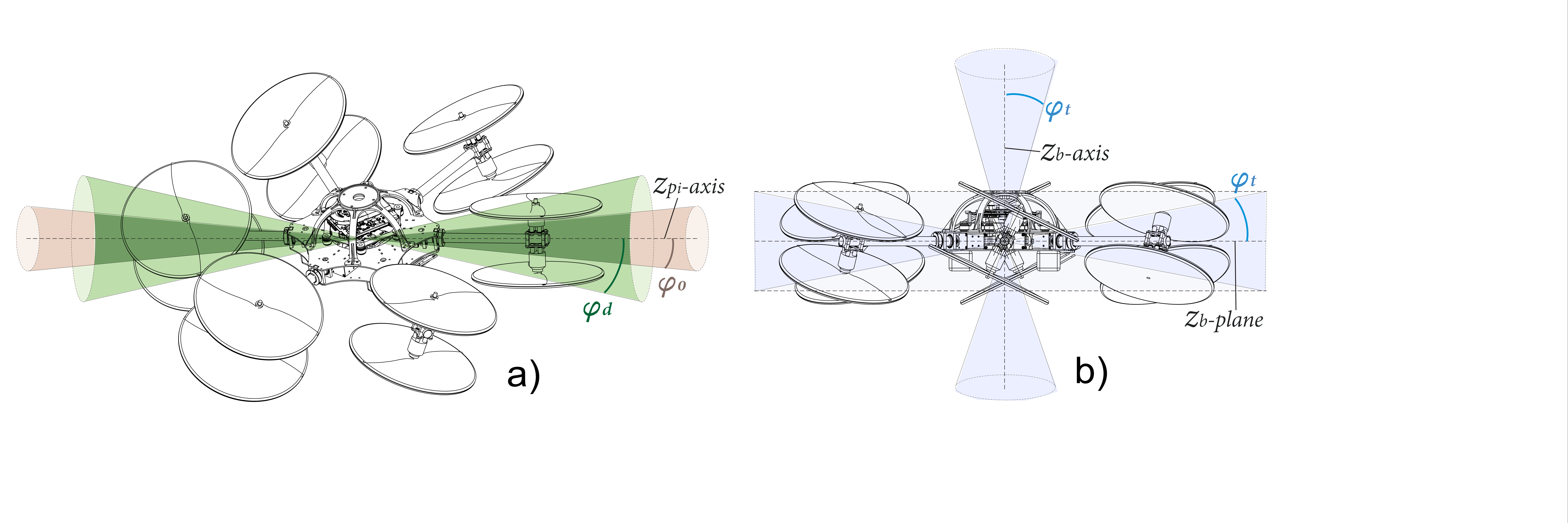}
 \end{subfigure}
 % ~
 \begin{subfigure}[b]{0.16\textwidth}
  \includegraphics[width=\textwidth]{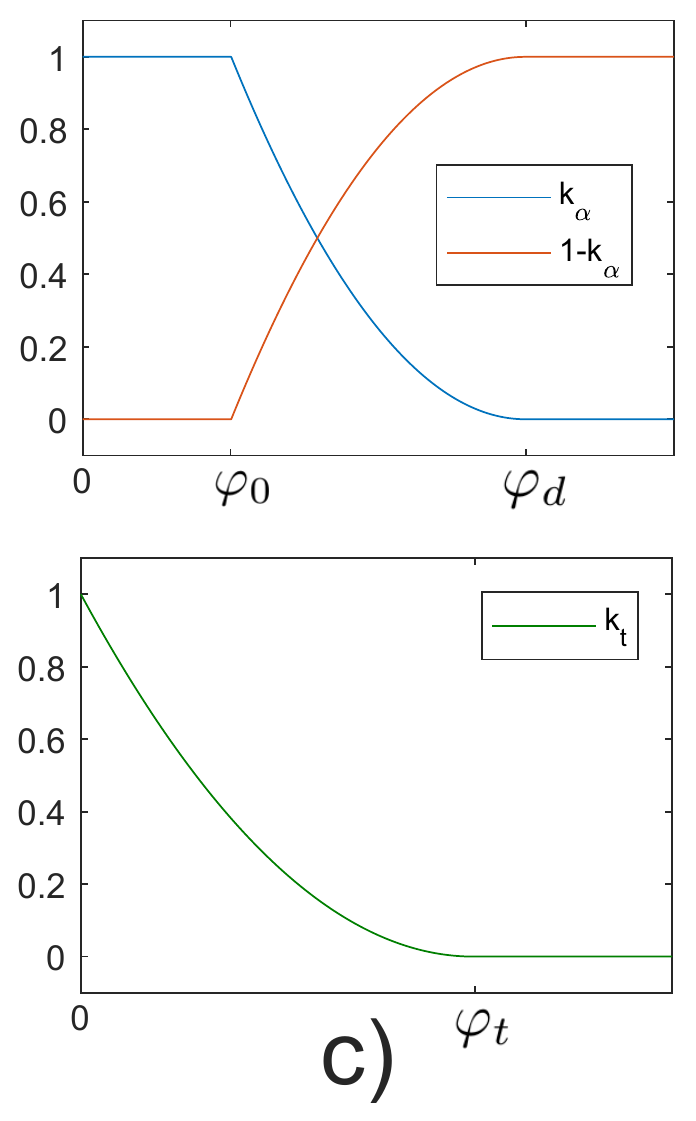}
 \end{subfigure}
 \caption{Visual depiction singularity handling for a) tilt angle singularities, b) kinematic singularties, and c) associated multipliers $k_\alpha$ and $k_t$.}\label{fig:singularity_handling}
\end{figure}

Omnidirectional wrench generation for the described vehicle encounters two types of singularity cases.
The first case is where alignment of the propellers results in a rank reduction of the instantaneous allocation matrix $\bm{A}_{\alpha}$, which corresponds to an ill-conditioned matrix (or high condition number).
This rank reduction occurs when the desired force is either orthogonal to the body $z_b$-frame (rank is reduced to 4, no instantaneous force exertion along $x_b$ or $y_b$ axes is possible),
and when the desired force lies in the $z_b$-plane (rank is reduced to 5, no instantaneous force exertion is possible along $z_b$). This phenomenon is shown in \autoref{fig:ForceCondition}(b), where the condition number asymptotically extends along the $z_b$-axis and $z_b$-plane.

The above mentioned kinematic singularities can be handled by identifying the case where the desired force vector comes within an angle $\varphi_t$ of the $z_b$-axis or the $z_b$-plane. A multiplier $k_t$ then quadratically scales an additional tilt component up to a value of $c_t$ when the misalignment angle $\phi$ goes to zero. Bias tilt directions $b_i$ alternate for neighboring tilt-arms. The modified tilt angle $\tilde{\alpha}_i$ can be expressed as

\begin{equation*}
    \begin{matrix}
  \tilde{\alpha}_i = \alpha_{i,prev} + \text{sign}(\delta\tilde{\alpha}_i)\text{min}(\vert\delta\tilde{\alpha}_i\vert, \dot{\alpha}_{max}\delta t)
  & \quad &
  \delta\tilde{\alpha}_i = \delta\alpha_i + k_{t} b_i c_t
\end{matrix}
\end{equation*}

\begin{equation*}
  k_{t} = \left\{
    \begin{matrix}
      0 & \forall & \phi \geq \varphi_{t} \\
      (1 - \frac{\phi}{\varphi_{t}})^2 & \forall & \phi < \varphi_{t}
    \end{matrix} \right.
     \quad \quad
    b_i = (-1)^{i} \forall i \in \{1...6\}
\end{equation*}

The resulting condition number envelope for the modified allocation matrix $\bm{A}_\alpha(\tilde{\alpha})$ is significantly reduced in the singularity cases, as seen when comparing \autoref{fig:ForceCondition}(b) and (c), where the scale of (c) is magnified for a better view of the new condition number envelope.
\autoref{fig:singularity_handling}(b) shows a visual interpretation of the thresholding approach described above, with the quadratic tilt bias multiplier $k_t$ shown in \autoref{fig:singularity_handling}(c).

The second singularity case is when the force aligns with an arm axis, causing a kinematic singularity where the associated thrusters cannot contribute to the desired force.
Since tilt angle dynamic limitations are not considered in the allocation, when the desired force crosses this axis directly,
it results in an angular velocity that approaches infinity as the time step goes to zero.
To expose the tilt motor angular velocities and evaluate this singularity condition, the allocation is taken to a derivative level, as proposed by Ryll et al.
To do this, tilt angle velocities are first expressed in the body wrench expression by including a structural null matrix:

\begin{equation}
 \begin{bmatrix}\bm{F}_{b} \\ \bm{\tau}_{b}\end{bmatrix}
 = \begin{bmatrix}\bm{A}_{\alpha}(\bm{\alpha}) & \bm{0}\end{bmatrix} \begin{bmatrix}\bm{\Omega} \\ \dot{\bm{\alpha}} \end{bmatrix}
 \label{eq:f_b}
\end{equation}

Now, the differential of $\bm{A}_{\alpha}(\bm{\alpha})$ may be expressed by expanding the terms as
\begin{equation*}
 \begin{matrix}
  \bm{A}_{\alpha}(\bm{\alpha})\bm{\Omega} & = & \sum\limits_{i=1}^{12} \bm{a}_i(\bm{\alpha})\Omega_i
  \\
  \frac{\text{d}\bm{A}_{\alpha}(\bm{\alpha})\bm{\Omega}}{\text{d}t} & =
  & \bm{A}_{\alpha}(\bm{\alpha})\dot{\bm{\Omega}} + \sum\limits_{i=1}^{12} \frac{\partial{\bm{a}}_i(\bm{\alpha})}{\partial\bm{\alpha}}\Omega_i\dot{\bm{\alpha}}

 \end{matrix}
\end{equation*}

\noindent and expressing the time derivative of (\ref{eq:f_b}) in the body frame as

\begin{equation}
 \begin{bmatrix}\dot{\bm{F}}_{b} \\ \dot{\bm{\tau}}_{b}\end{bmatrix}
 = \begin{bmatrix}\bm{A}_{\alpha}(\bm{\alpha}) & \sum\limits_{i=1}^{12} \frac{\partial{\bm{a}}_i(\bm{\alpha})}{\partial\bm{\alpha}}\Omega_i\end{bmatrix}
 \begin{bmatrix}\dot{\bm{\Omega}} \\ \dot{\bm{\alpha}} \end{bmatrix}
   \label{eq:d_fb}
\end{equation}

From expression (\ref{eq:d_fb}) it is clear that the system experiences a singular condition in the tilt angle velocities when an $\Omega_i$ term is zero,
which occurs when the force and torque capabilities of the given propeller unit cannot contribute to the desired wrench.
This corresponds with a case where the desired force and torque vectors align with the propeller group $x_{p_i}$-axis.
An approach commonly used in robot manipulators can be invoked to reduce this issue, by instead using the damped pseudo-inverse of the allocation matrix \cite{wampler1986manipulator}.
This approach, however, requires the inversion of a 6x6 matrix in real time, which currently due to high use of CPU and memory make it a poor choice for use on a micro-controller.
Taking inspiration from this approach, a solution that approximates a damping effect is derived, requiring significantly less computational power.

When the angle between the desired force vector comes within the damping threshold angle $\varphi_{d}$ of an arm axis,
the tilt-angle speeds associated with that axis are quadratically scaled to zero by multiplier $k_{\alpha,i}$.
As the alignment angle $\eta_i$ approaches a zero threshold, $\varphi_{0}$,
the tilt angle freezes in position.
A visual interpretation of the alignment geometry is shown in \autoref{fig:singularity_handling}(a), with multiplier curves depicted in \autoref{fig:singularity_handling}(c).
The change in $\alpha_i$ can then be expressed as follows:

\begin{equation*}
 \delta \alpha_i ' = \delta \alpha_i (1 - k_{\alpha,i})
\end{equation*}
\begin{equation*}
 k_{\alpha,i} = \left\{
 \begin{matrix}
  0                                                          & \forall & \eta_i > \varphi_{d}               \\
  (1 - \frac{\eta_i - \varphi_{0}}{\varphi_{d} - \varphi_{0}})^2 & \forall & \varphi_{d} > \eta_i > \varphi_{0} \\
  1                                                          & \forall & \eta_i < \varphi_{0}
 \end{matrix} \right.
\end{equation*}

This approach is justified due to the fact that at small angles of alignment deviation,
the force contribution of the thrusters in question is negligible when compared to the remaining thrust contributions, and the error can be treated as a disturbance.
When the new angles are computed based on these values, and limited according to $\dot{\alpha}_{max}$,
the rotor speeds are then calculated as in (\ref{eq:omegas}).
A force from the limited propeller group exists only if the thrust axis can be projected as a positive component onto the desired force vector.

Since the contribution of propeller groups in the tilt-angle singularity region is so small, one can take advantage of this situation to address another issue of the physical system, notably the windup of cables around an axis.
As the tilt velocity is ramped down, it is balanced with an unwinding velocity $\omega_u$, that is calculated based on a maximum unwinding rate and direction to the zero position,
 resulting the new tilt angle change $\delta \alpha_i^* = \delta \alpha_i' - \text{sign}(\alpha_i) k_{\alpha,i} \omega_u \delta t$.

\section{Experimental Results}
The presented results are derived from flight tests in an indoor space, using a VICON external camera system for state estimation.
Parameter values used for the experiments are shown in \autoref{tab:values}.

\begin{table}
\centering
\begin{tabular}{c|c|c|c|c}
\hline
$\varphi_{0} [\text{deg}]$ & $\varphi_{d} [\text{deg}]$ & $\varphi_{t} [\text{deg}]$ & $c_{t} [\text{deg}]$ & $\omega_{u} [\frac{\text{rad}}{s}]$ \\
\hline \hline
5 & 15 & 10 & 10 & 8 \\
\hline
\end{tabular}
\caption{Parameters used for experimental flight tests}
\label{tab:values}
\end{table}

\begin{figure}[ht!]
 \centering
 \includegraphics[width=125mm]{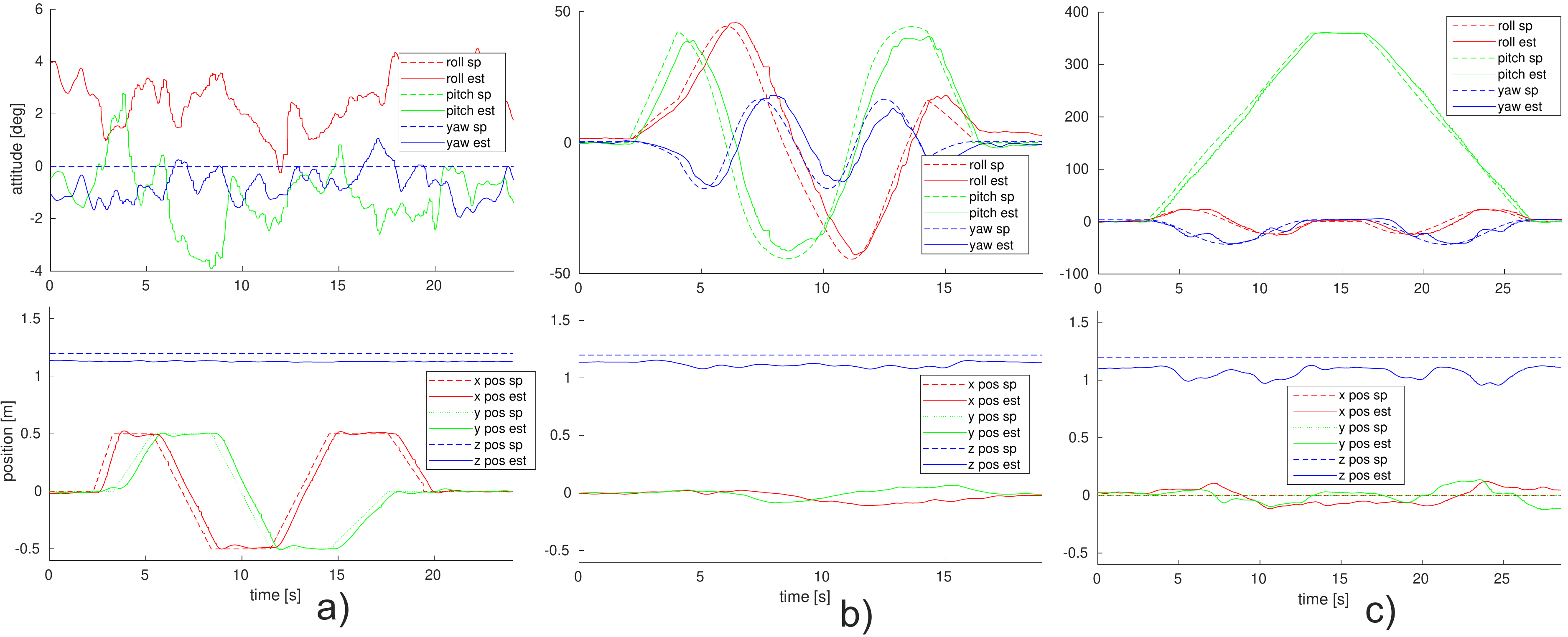}
 \caption{Experimental trajectory tracking a) translation without rotation, b) rotation without translation c) 360 degree flip about the $y_b$-axis}
 \label{fig:decoupling}
\end{figure}

\paragraph{Omnidirectional flight:}
To validate the controller, three trajectories were tested to demonstrate decoupling of the force and torque. Tracking results are shown in \autoref{fig:decoupling}. The first two trajectories demonstrate the decoupled translation and rotation of the system, and the third trajectory is a 360$^\circ$ flip about the $\bm{y}_b$-axis. The system is able to track these trajectories with position errors below 5 cm and attitude errors below 4$^\circ$.

\begin{figure}[ht!]
 \centering
 \includegraphics[width=125mm]{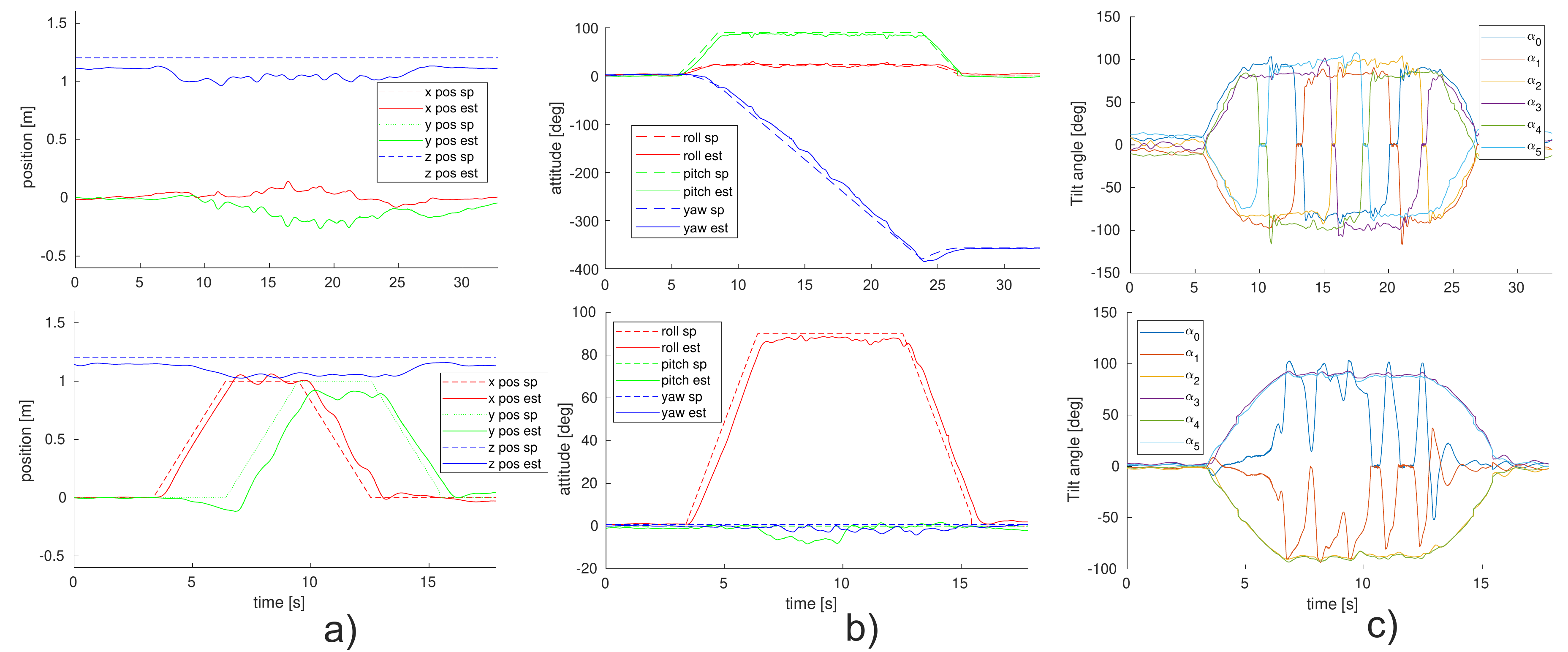}
 \caption{Experimental a) position tracking, b) attitude tracking, and c) commanded tilt angles for translation in a singular configuration (upper row) and a cartwheel through 6 singular positions (lower row)}
 \label{fig:singularity}
\end{figure}

\paragraph{Robustness to singular configuration:}
In order to demonstrate robustness to the singularity cases presented above, two additional trajectory tracking experiments have been conducted, with results shown in \autoref{fig:singularity}. The first trajectory rotates the vehicle such that one arm is aligned with gravity, then tracks a translation trajectory.
The second trajectory moves the vehicle to 90$^\circ$ pitch, then completes a 360$^\circ$ rotation about the body frame $\bm{z}$-axis.
In both cases which pass through known singularity regions, experimental results confirm stable tracking of the desired trajectory.

\begin{figure}
 \centering
 \includegraphics[width=125mm]{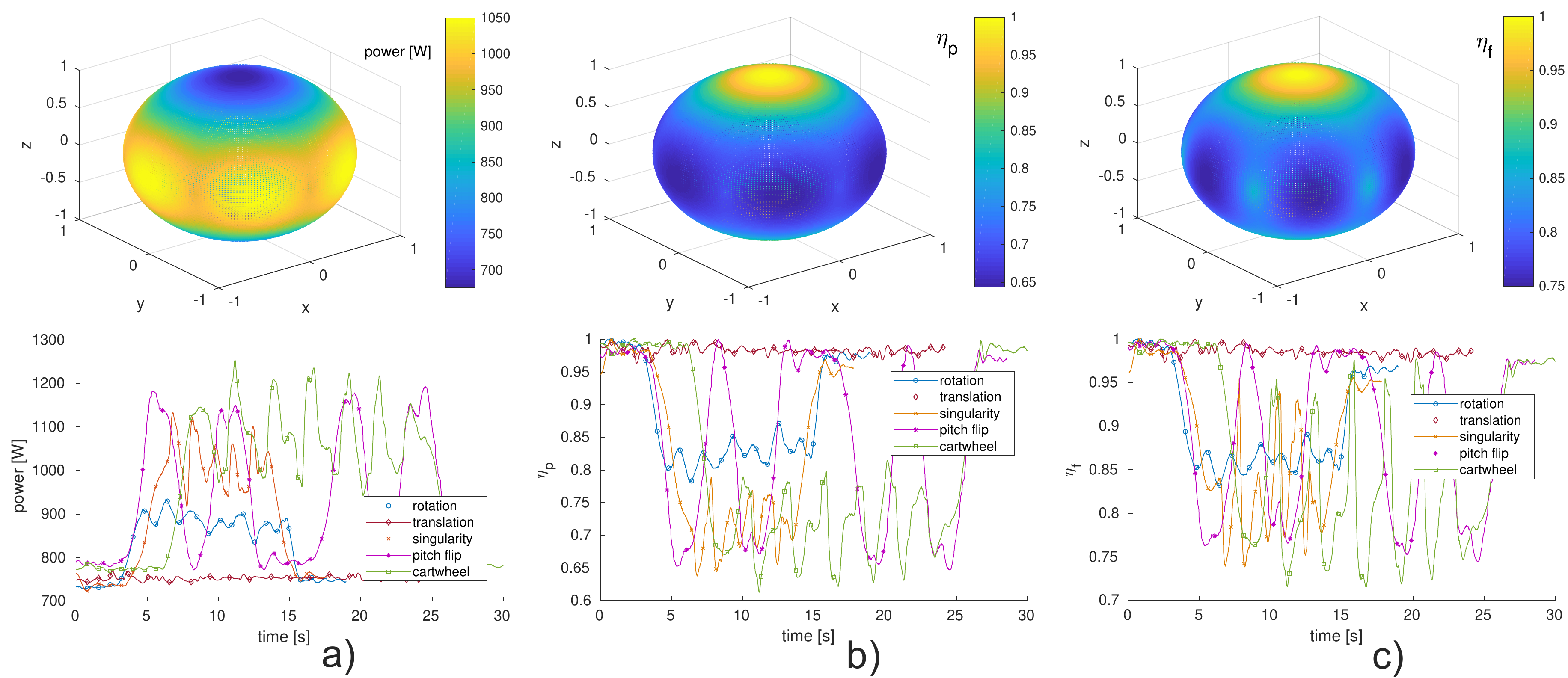}
 \caption{a) Power consumption, b) power efficiency and c) wasted force index for theoretical hover, and measured values from tested trajectories.}
 \label{fig:efficiency}
\end{figure}

\paragraph{Efficiency Evaluation:}
Efficiency is defined here in terms of relative power consumption, and evaluated relative to the best case power consumption $p_h$ of horizontal hover,
$\eta_P = \frac{v_{total}\Sigma_{i=1}^{12}{i_i}}{p_{h}} \in [0,1]$.
A wasted force index is also computed, as proposed in \cite{ryll2016modeling},
$\eta_f = \frac{{\lVert}\bm{F}_b{\rVert}}{\Sigma_{i=1}^{12} f_i} \in [0,1]$, where $f_i$ are the thrust values for each rotor.
Estimated total power, power efficiency and wasted force for each hover configuration are shown in \autoref{fig:efficiency}, projected onto the force envelope.
Below the theoretical envelopes, measured values from experimental results are displayed, using measured PWM-to-current correlation and flight-test voltage logs to find the power consumed.
Experimental results agree well with the ranges predicted on the theoretical envelopes, showing that efficient flight configurations are achieved in specific orientations, while the system remains pose-omnidirectional with varying degrees of efficiency.

\section{Conclusion}
\vspace{-1.5mm}
Results presented in this paper demonstrate that although actuated tilting rotor units have added complexity and weight compared to other fully actuated MAVs, the additional weight can be justified by improved hover efficiency of the platform in particular configurations, and additional complexity by the reliability of the prototype system, VoliroX.

As a result of the assumption that rotor thrust and drag torques are independent, any airflow interference between rotor groups is treated as an unknown external disturbance to be rejected. This is a reasonable assumption for some but not all body orientations, such as when flying the cartwheel maneuver, where lower rotors are affected by airflow from those above. As a result there is some loss in tracking performance, solutions to which are being developed in ongoing work.
Moving to more dynamic flight trajectories, Further techniques for singularity avoidance will be investigated, and additional benefits that can be yielded from an overactuated system.

\vspace{-1.5mm}
\paragraph*{Acknowledgements: }
This work was supported by funding from ETH Research Grants, the National Center of Competence in Research (NCCR) on Digital Fabrication, NCCR Robotics, and Armasuisse Science and Technology.

\vspace{-1.5mm}
\bibliographystyle{ieeetr}
\bibliography{references}

\begin{thebibliography}{10}

\bibitem{brescianini2016design}
D.~Brescianini and R.~D'Andrea, ``Design, modeling and control of an
  omni-directional aerial vehicle,'' in {\em ICRA 2016}, pp.~3261--3266, IEEE,
  2016.

\bibitem{park2018odar}
S.~Park, J.~Lee, J.~Ahn, M.~Kim, J.~Her, G.-H. Yang, and D.~Lee, ``Odar: Aerial
  manipulation platform enabling omni-directional wrench generation,'' {\em
  submitted to IEEE/ASME Transactions on Mechatronics}, 2018.

\bibitem{park2016design}
S.~Park, J.~Her, J.~Kim, and D.~Lee, ``Design, modeling and control of
  omni-directional aerial robot,'' in {\em IROS 2016}, pp.~1570--1575, IEEE,
  2016.

\bibitem{staub2018towards}
N.~Staub, D.~Bicego, Q.~Sabl{\'e}, V.~Arellano, S.~Mishra, and A.~Franchi,
  ``Towards a flying assistant paradigm: the othex,'' in {\em ICRA 2018}, IEEE,
  2018.

\bibitem{falconi2012dynamic}
R.~Falconi and C.~Melchiorri, ``Dynamic model and control of an over-actuated
  quadrotor uav,'' {\em IFAC Proceedings Volumes}, vol.~45, no.~22,
  pp.~192--197, 2012.

\bibitem{ryll2015novel}
M.~Ryll, H.~H. B{\"u}lthoff, and P.~R. Giordano, ``A novel overactuated
  quadrotor unmanned aerial vehicle: Modeling, control, and experimental
  validation,'' {\em IEEE Transactions on Control Systems Technology}, vol.~23,
  no.~2, pp.~540--556, 2015.

\bibitem{kamel2018voliro}
M.~Kamel, S.~Verling, O.~Elkhatib, C.~Sprecher, P.~Wulkop, Z.~Taylor,
  R.~Siegwart, and I.~Gilitschenski, ``Voliro: An omnidirectional hexacopter
  with tiltable rotors,'' {\em arXiv preprint arXiv:1801.04581}, 2018.

\bibitem{ryll2016modeling}
M.~Ryll, D.~Bicego, and A.~Franchi, ``Modeling and control of fast-hex: a
  fully-actuated by synchronized-tilting hexarotor,'' in {\em IROS 2016},
  pp.~1689--1694, IEEE, 2016.

\bibitem{lee2010geometric}
T.~Lee, M.~Leoky, and N.~H. McClamroch, ``Geometric tracking control of a
  quadrotor uav on se(3),'' in {\em Decision and Control 2010}, pp.~5420--5425,
  IEEE, 2010.

\bibitem{invernizzi2017geometric}
D.~Invernizzi and M.~Lovera, ``Geometric tracking control of a quadcopter
  tiltrotor uav,'' {\em IFAC-PapersOnLine}, vol.~50, no.~1, pp.~11565--11570,
  2017.

\bibitem{wampler1986manipulator}
C.~W. Wampler, ``Manipulator inverse kinematic solutions based on vector
  formulations and damped least-squares methods,'' {\em IEEE Transactions on
  Systems, Man, and Cybernetics}, vol.~16, no.~1, pp.~93--101, 1986.

\end{thebibliography}
\end{document}